\documentclass{apa7}

\usepackage{lipsum}
\usepackage[american]{babel}
\usepackage{graphicx}
\usepackage{placeins}
\usepackage{setspace}

\usepackage{lscape}
\usepackage{longtable}
\usepackage{booktabs}
\usepackage{array}
\usepackage{ragged2e}
\usepackage{makecell}
\usepackage{amsmath}

\usepackage{appendix}
\usepackage{listings}
\usepackage{xcolor}
\usepackage{tabularx}
\usepackage{forest} 
\usepackage{xcolor} 
\usepackage{pdflscape}
\usepackage{rotating}

\usepackage{csquotes}
\usepackage[style=apa,sortcites=true,sorting=nyt,backend=biber]{biblatex}
\DeclareLanguageMapping{american}{american-apa}
\title{Discourse Analysis of Legislative Bill Discussions}
\shorttitle{Discourse Analysis}

\author{Isaac Rudnick, Nipun Das, Foaad Khosmood}
\affiliation{California Polytechnic State University}

\leftheader{Discourse}

\abstract{Legislative hearings are distinct communication events with their own human recognizable structure and logic, yet they have been analyzed as distinct phenomena beyond their transcripts. Understanding legislative hearings at a deeper level is a major step in analyzing the work and power relationships within democratic bodies. In this work, we present a novel end-to-end pipeline for parsing rhetorical structures from legislative committee hearings with the goal of improving legislative discourse analysis and democratic accountability. We also propose a rhetorical structure grammar and find that 74.71\% of California state bill discussions in our dataset can be parsed with this grammar. We evaluate our pipeline with two utterance section classifiers on a random subset of over 2619 annotated utterances corresponding to 100 hearings. We find that a logistic regression-based classifier with grammar-based output masking and utterance tag smoothing outperforms more complex models, achieving an F1 score of 59.0\% against our ground truth discourse section labels.}

\keywords{NLP, Discourse Representation Theory, Parsing, Digital Humanities}

\authornote{
  Correspondence concerning this article should be addressed to Foaad Khosmood, Department of Computer Science and Software Engineering, California Polytechnic State University, 1 Grand Av., San Luis Obispo, CA 93401.  E-mail: foaad@calpoly.edu
}

\begin{document}
\maketitle

\section{Introduction}
In most representative systems such as the US Federal government and US State governments, much of the work is done in legislative committees. Committees represent an important step in the legislative process. They are formed around policy topics and their main function is to hear, evaluate and vote on bills. Much of this work takes place during committee hearings in which the contents of one or more bills are discussed. In most US state governments, the main part of these hearings can be divided into ``bill discussions'' where each discussion concerns a distinct bill.

Legislative bill discussions themselves are often composed of multiple stages. These may include introductions, bill presentations, testimony from expert witnesses, comments from the general public, and voting. Creating a pipeline to parse legislative bill discussions according to these stages makes it possible to label numerous existing legislative bill discussions, such as those collected by the OpenStates and Digital Democracy projects \parencite{Khosmood2024} \parencite{Plural2023}. This, in turn, allows viewers of these transcripts to better understand the different sections of a bill hearing and thus the transcript overall. 
A clean, standardized representation also makes legislative transcripts and resources more broadly available. Data aggregated by the OpenStates project, for example, has seen significant use by the American Civil Liberties Union (ACLU), NPR's StateImpact, and various journals \parencite{Plural2023}. 
Data availability and cleanliness are not enough, however, as extracting and interpreting insights from this data still requires significant processing.

While often performed manually at present, analysis of the discourse patterns in bill discussions can help derive insights into the general structure of these discussions and how some bill discussions vary from others. An automated approach therefore creates significant opportunity for broader civic engagement.

Given a bill discussion in the form of transcribed units of speech, the problem of discourse parsing can be formalized as a classification problem where each unit of speech is labeled with one or more tags that represent the (sub)sections of the structure of the bill discussion. This allows the overall transcript to be parsed into sections that represent the different stages of the discussion, and the labels can be used to derive a grammar for bill discussions in general.

For the Digital Democracy dataset, bill discussion transcripts are provided as a series of utterances. An utterance is a unit of speech by a speaker, and each utterance record within the dataset includes the speaker's id number (\texttt{pid}) and the transcribed text spoken by the speaker \parencite{Khosmood2024}.

Starting with a hearing event sequence schema, we build a data model using a variant of the process described by \textcite{Rudnicki2016}. A bottom-up approach with only n-grams and the original schema we were provided might still miss the low-hanging fruit of speakers' \texttt{pid}s. This requires basic named entity recognition and/or \texttt{pid} resolution. To take advantage of this and ensure broader consistency, we plan to tag named entities associated with the top-level ontological terms used by the UK parliament taxonomy \parencite{Somerville}. Tagging and abstracting away the associated named entity references in this way should help with n-gram extraction, as speakers' roles are often more contextually relevant than particular individuals' names or identities.

To associate n-grams with their relevant sections, we use a combination of manual and automated filtering. For the automated filtering process, we filter n-grams based on their parent utterances' relative positions within the committee hearing transcripts. By only reviewing token sequences that frequently occur towards the beginning, middle, and end of a bill discussion, we can narrow down a set of key-phrases for section tagging. Finally, a Beginning-Inside-Outside (BIO) tagging system is used to construct the different sections' spans.

Manual review of the most frequent and highest co-occurring key phrases will be used in the absence of a better method. This manual review allows us to gain insight into which phrases are indicative of transitions between sections, which we used to create regular expression and fuzzy matching rule-based transition detectors. We utilize the patterns observed from this manual review during feature extraction to create features indicative of section transitions.

On the task of tagging of outer utterance sections, we achieve a macro average precision of 58.7\%, recall of 62.4\%, accuracy of 70.1\%, and an F1-score of 59.0\% using a grammar-masked logistic regression (LR) classifier. Evaluation was conducted against a validation set of LLM and human-tagged ground-truth utterance section labels. This performance was especially interesting, as the masked LR model outperformed a larger, unmasked model on the same task by 8.0\% in F1 and 3.3\% in accuracy.

In the remainder of this paper, we discuss the background of legislative hearings within government. We then discuss the dataset, related research in the field, our methodology, and the results of our work. We finish with the conclusions that can be drawn from our work and potential future work based on our research.

\section{Background}
To understand the methodology and approach used, it is useful to understand the general structure of legislative hearings. Governments legislatures at the state and federal levels consist of multiple legislators, and these legislators often are members of committees that focus on different topics. Committees consist of a subset of legislators, and these committees hold hearings where bills are discussed, amended, and voted on. Hearings consist of multiple bill discussions, and each bill discussion includes multiple potential stages \parencite{Khosmood2024}. These include introductions, witness testimony, comments from the general public, discussion among legislators, and voting.

We focus on a selection of what we consider the most relevant discourse information that was found to be missing from the dataset in particular speaker role information, including who is the presiding officer who can file different motions and procedures and differentiating expert testimonials from public testimonials.

In terms of discourse structure, hearing discussions are segmented into seven sections based on observation:
\begin{enumerate}
    \item \textbf{Introductory Section.} This section includes general opening procedures, such as the presiding chair calling for the Pledge of Allegiance or delivering another opening statement.

    \item \textbf{Bill Presentation.} This section includes a lay summary of the bill's contents, typically provided by the bill author or the presiding chair.

    \item \textbf{Committee Discussion.} This section includes discussion of the bill's contents among committee members and, when applicable, the bill presenter.

    \item \textbf{Expert Testimony.} This section includes testimony from experts invited or scheduled by committee members to discuss specialized knowledge related to the bill and respond to questions.

    \item \textbf{Public Testimony.} This section includes testimony from members of the public, distinct from experts and legislators, who express support for or opposition to the bill. Such testimony may be offered on behalf of an interest group, such as a teachers' association, steelworkers' union, or similar organization.

    \item \textbf{Closing Remarks.} This section includes closing statements delivered by the presiding chair or bill presenter, including final arguments in support of the bill or concluding descriptions of its purpose.

    \item \textbf{Voting Procedures.} This section includes procedural actions related to voting, such as seconding the bill, making a motion to pass, conducting a roll call, and completing related voting procedures.
\end{enumerate}

Three fallback classes (`OTHER\_HEARING', `OTHER\_PROCEDURAL', `OTHER\_NONPROCEDURAL') were assigned as appropriate according to the speaker role requirements depicted in Figure \ref{role associations} and by the grammar in Appendix \ref{grammar} when possible.

Through preliminary analysis and SME consultation, we determined that these sections frequently occur in an order described by the grammar given in Appendix \ref{grammar}.

\subsection{Dataset}\label{dataset}
To develop and evaluate our model, we use a dataset developed by the Institute for Advanced Technology and Public Policy as part of the ``Digital Democracy'' government transparency project.
The Digital Democracy dataset is a collection of various hearings, bill discussions, and named entity information associated with legislatures' hearing transcripts \parencite{Khosmood2024}.
The original corpus contains hearing transcript and bill data from California, Texas, New York and Florida for 2015-2016 (California only) and 2017-2018.
For this work, however, we annotate a random subset of hearing transcripts associated with the 2015-2016 and 2017-2018 California Senate and Assembly committee hearings. We leave out senate resolution discussions, which are largely procedural, and floor discussions that did not pertain to any particular bills. This filtering brought the total number of bills sampled down from 8218 to 7030.
Due to resource constraints, we knew we wouldn't be able to train and evaluate on the full dataset. On the other hand, testing on smaller sets with 20 and 50 annotated hearings often resulted in some section classes and speaker roles having 0 support during training or evaluation. We ultimately settle on a ground truth set of 2619 annotated utterances across 100 hearings. The unannotated collection of hearings' utterances, extracted features, and the SpeakerPositionEnum and SectionEnum definitions were provided to ChatGPT-5.5 for a first round of labelling. Manual inspection found these annotations to be lacking, so Isaac manually annotated the first 34 hearings' utterances. This annotated subsample was fed to Anthropic's Fable model and several rounds of manual inspection and corrections were performed.

\section{Related Works}
\subsection{Legislative Ontologies}
Due to the stakes surrounding their availability and usability, various nations have dedicated time and resources to the standardization of legal texts. One of the leaders in this area of standards is the United Kingdom's parliament. Beginning in 2016, the UK Parliamentary Digital Service team began constructing various ontologies for different procedures, resources, materials, and entities within the UK parliament. This was primarily done to improve internal and external accessibility and redesign the UK parliament's Content Management System (CMS) for interoperability \parencite{Smethurst2016} but quickly developed into a broader attempt to formalize the parliamentary ecosystem \parencite{Somerville}. While the project lacks natural language parsing tools, their ontologies and data models still present extremely useful schemata and guidance on segmenting broader legislative discussions \parencite{Somerville}.

A separate, independently-run project based in the United States called OpenStates began with a similar goal: ``improve civic engagement by providing data and tools to help you understand the legislative process that creates public policy'' \parencite{Plural2023}. It began in 2009 and primarily acts as a national legislative data aggregation, standardization, cleaning, and publishing organization for bulk downloading. Since then, a small collection of volunteers has constructed a knowledge graph-based data model for tracking and standardizing hearings \parencite{Turk2017}. This has seen a large amount of use by the ACLU, NPR, and various journals \parencite{Plural2023}. While not as detailed as the UK parliamentary ontologies, they represented a very useful resource when building our own data model.

\subsection{Discourse Parsing}
The field of discourse parsing has exploded since Propp's original work on folklore narrative structures. In their 2022 literature review, Li, Liu, Qin, and Liu compare several different discourse parsing techniques \parencite{Li2022}. They describe three main tasks within the general research area of discourse parsing, namely Rhetorical Structure Theory (RST), Penn Treebank (PDTB), and discourse parsing for multiparty dialogue.

Rhetorical Structure Theory (RST) aims to parse documents into a tree structure. The document is segmented into elementary discourse units (EDUs), which make up the leaves of this tree, and these leaf nodes are connected through relations. Various models have been proposed for extracting these parse trees, but most recent successes have used neural models \parencite{Li2022} \parencite{Wang2018} \parencite{Knaebel2021}. In their 2018 work, for example, \textcite{Wang2018} develop an end-to-end neural segmentation pipeline that achieved state-of-the-art (SotA) performance on the RST-DT dataset. To do this, the authors fine-tuned a larger model with restricted self-attention, giving an F1 score of 94.3\% at significantly lower processing time \parencite{Wang2018}.

The Penn Discourse Tree Bank (PDBT) is a dataset that represents relationships between arguments using either explicit or implicit discourse relations. It was developed to provide a ground-truth set of lexically-grounded discourse relation annotations. Each annotation includes discourse connective and pairwise argument information \parencite{Miltsakaki2004}. It has long been a staple for benchmarking discourse annotation approaches \parencite{Li2022}. Despite its usefulness and breadth, however, the project remains focused on annotating the  Wall Street Journal (WSJ) section of Treebank-2 \parencite{Miltsakaki2004}.
A recent work by \textcite{Knaebel2021} introduced a modular, extensible neural pipeline for shallow discourse parsing that they term \texttt{discopy}. \texttt{discopy} achieves modest F1 scores (72.16\% to 88.68\%) on the PDTB dataset for thresholded partial-overlap argument tagging \parencite{Knaebel2021}. This pipeline is evaluated with various BERT embedding models with sliding windows for token-level argument segmentation.
Unfortunately, the structure of discourse structures in this corpus thus makes the development of multi-party dialogue discourse parsers more difficult. Thus, discourse parsers of multi-party dialogues often rely on hand-crafted features, used with either "shallow" machine learning models or "deep sequential models" \parencite{Li2022}. As a unique benefit of being open-source, \texttt{discopy} still holds promise for efficient multi-party dialog parsing \parencite{Knaebel2021}.

Each of these styles of discourse parsing have various implementation methods, such as supervised machine learning models that label the boundaries of elementary discourse units (EDUs) and maximum spanning tree algorithms using features like punctuation and position for classifying EDUs in multi-party dialogue.
Ultimately, our legislative bill discussion parser does not require segmentation into EDUs, since the dataset we work with already provides the discourse segmented by utterance. Also, the end goal of our discourse parser is also not to generate a parse tree or labels at the EDU level, but rather at the utterance level, understanding the general structure of bill discussions and the transitions between broad sections rather than the structures of each utterance itself.
For this work, then, we took inspiration from the restricted self-attention approach used by \textcite{Wang2018} and opted for RST parse tree-based masking.

The unique structure of our dataset also motivated a shallow, hand-crafted feature approach for preliminary utterance tagging. We find that manual exploration of our legislative bill discussion dataset and hand-crafted features based on common patterns or key phrases was a key contributor to our model's effectiveness.

\subsection{Digital Democracy Transcript Analysis}
The Digital Democracy dataset has been used in a variety of NLP research, including tasks adjacent to transcript discourse parsing. In their 2026 works, \textcite{Dimitrijevik2026} and \textcite{Das2026} took different approaches to tagging self-introductions with transcripts of the Digital Democracy corpus. We extend this work to parsing other sections of our discourse. After segmenting the overall bill transcript into various sections, our parser has the opportunity to further parse bill discussions through tasks such as counting votes, identifying the identity, organization, and stance of public testimony witnesses, and assigning roles specific to bill discussions, like bill presenter.

\subsection{Legislative Debate Discourse Analysis}
In a work by \textcite{Irani}, a framework called DALiSM is introduced for analysis of legislative debates, such as committee hearings. The generalized framework is applied both to U.S. congressional hearings as well as social media debates on the Reddit platform. The authors of \textcite{Irani} introduce an ``argument-centric'' approach, using various machine learning and large language models to identify arguments and track the discourse of arguments in legislative hearings or on social media platforms. Although the output of the DALiSM framework differs from the utterance level  section tags that we are providing, the methodology from this paper provides useful insight into some of the existing machine learning models and technical approaches that can be applied to political discourse analysis.

\section{Methods}
\subsection{Exploratory Data Analysis}
\begin{figure*}[t]
    \centering
    \includegraphics[width=1.00\textwidth]{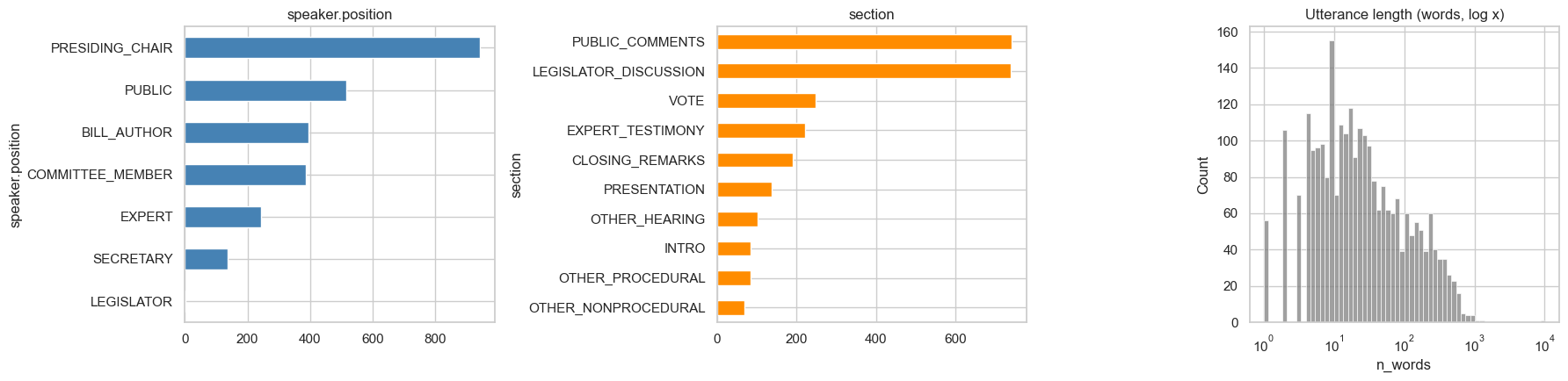}
    \caption{Analysis of the ground-truth labeled sample showed the following occurrence frequencies for different speaker roles, section types, and utterance lengths. The outlier utterance at around $10^4$ tokens was due to a segmentation error for bill `CA\_201720180AB10`, hearing `52708`.}
    \label{speaker_section_frequencies}
\end{figure*}

Exploratory analysis revealed 231 California hearings with unidentified speakers. We found that the majority of such errors were caused by transcription errors, although a few referenced IDs not associated with any individual in the California dataset.
Additional summary EDA information can be found in \ref{speaker_section_frequencies}

\subsubsection{Utterance Normalization}
Our pipeline begins by loading bill discussions into Python data classes from the raw CSVs of the Digital Democracy dataset. A combination of fuzzy matching, regular expressions, and Named Entity Recognition (NER) performed by SpaCy's `en\_core\_web\_sm` model were used to perform basic coreference resolution. Tagged mentions are mapped to common identifiers (such as `BILL') for more complex textual feature extraction. These token substitutions also enhance later TF-IDF-based ngram embedding by simplifying token sequences. Referent types and their associated keyphrase tokens/replacement strings are described in Table \ref{ref_replacement_tokens}. These simplified token representations helped with more complex feature generation downstream and TF-IDF embedding analysis. 

\begin{table}[htbp]
\centering
\caption{Different types of named entities' in-text mentions were replaced with more abstract, 1-token strings described in the `Replacement' column. Examples of mentions that would be replaced are shown in the `Example' column.}
\label{ref_replacement_tokens}

\begin{tabular}{p{0.10\textwidth} p{0.15\textwidth} p{0.2\textwidth}}
\hline
\textbf{Referent Type} & \textbf{Example} & \textbf{Replacement} \\
\hline
Speaker &
Debra Bowen &
\texttt{SpeakerPositionEnum} associated with the individual referenced \\
\hline
Bill &
AB 420&
`BILL' \\
\hline
Organization &
California Polytechnic State University &
`ORG' \\

\hline
\end{tabular}
\end{table}

A common regex pattern we identify is the bill identifier (ex. ``AB 743''). The bill identifier is not commonly said during the discussion, typically only by the bill presenter and occasionally by the chairperson when introducing the presenter and/or the committee secretary when calling the vote. To account for this, we match against common phrases and terms used to reference such entities. A similar approach is used for other referents.

After this coreference resolution and token sequence substitution process, we normalize all non-special tokens by removing punctuation and casting them to lowercase.

\subsubsection{Ngram Extraction and Analysis}
After this text normalization process, we collect term and document frequencies associated with token ngrams of length 2 through 9.
To filter out redundant ngrams, we dropped any ngrams that represented subsequences of longer ngrams with a co-occurence threshold of $\geq $60\%. For example, if the token ``thank'' co-occurred with bigram ``thank you'' 90\% of the time, then only the ``thank you'' bigram would be kept for feature analysis. In addition to this maximal ngram-based filtering, we only kept ngrams that appeared in at least 5\% of the sample dataset's utterances. Ngram feature count by sequence length is shown in Table \ref{tab:ngram keep drop rates}.

\begin{table}[ht]
\centering
\caption{Rates at which shorter token ngrams were dropped in favor of longer ones for TF-IDF vectorization.}
\label{tab:ngram keep drop rates}
\begin{tabular}{p{0.05\textwidth} p{0.05\textwidth} p{0.05\textwidth} p{0.1\textwidth} p{0.1\textwidth}}
\hline
\textbf{Order} & \textbf{Total} & \textbf{Kept} & \textbf{Dropped} & \textbf{Dropped (\%)} \\
\hline
2 & 9890 & 8971 & 919 & 9.3 \\
\hline
3 & 5858 & 5208 & 650 & 11.1 \\
\hline
4 & 2216 & 1894 & 322 & 14.5 \\
\hline
5 & 950 & 775 & 175 & 18.4 \\
\hline
6 & 455 & 366 & 89 & 19.6 \\
\hline
7 & 239 & 198 & 41 & 17.2 \\
\hline
8 & 128 & 115 & 13 & 10.2 \\
\hline
9 & 71 & 71 & 0 & 0.0 \\
\hline
\end{tabular}
\end{table}

\begin{figure*}[t]
    \centering
    \includegraphics[width=\textwidth]{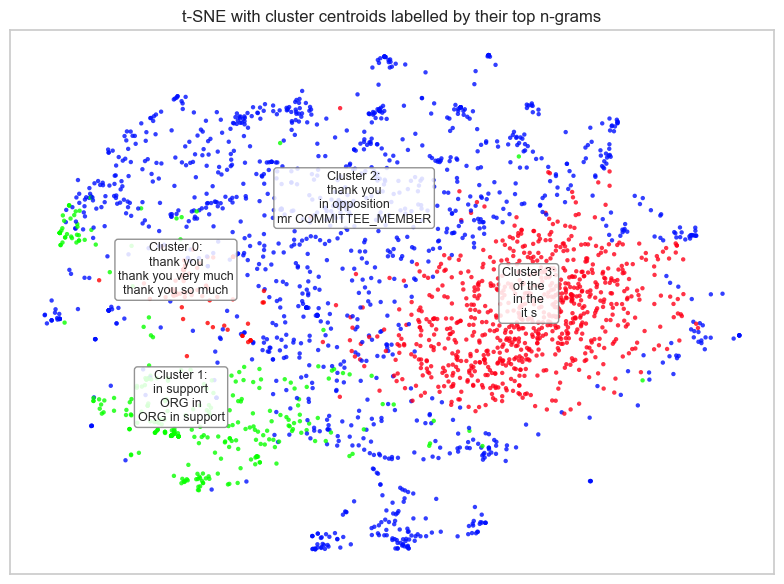}
    \caption{T-SNE clusters associated with the dimensionality-reduced ngram TF-IDF embeddings.}
    \label{t-sne clusters}\label{fig:confusion_matrix}\label{ngram correlations}\label{grammar}
\end{figure*}

To perform additional analysis, we created TF-IDF embeddings for this consolidated feature set.
Elbow analysis of the silhouette scores associated with different k-means cluster counts for the TF-IDF embeddings revealed 4 clusters to be the best balance of cluster count and cluster fit. That said, the silhouette score was only about 0.087. A visualization of these clusters using T-SNE for dimensionality reduction can be seen in Figure \ref{t-sne clusters}.

The Pearson R's of the top-10 most highly correlated ngrams for each speaker and section can be seen in Figures \ref{speaker_ngrams} and \ref{section_ngrams} in Appendix \ref{ngram correlations}.

\subsection{Data Model}
Due to the size and complexity of the UK Parliament ontologies, we relied primarily on the feedback of Subject Matter Experts (SMEs) and related works in legislative discourse parsing for developing a unified speaker, utterance, and role ontology \parencite{Dimitrijevik2026} \parencite{Das2026}. We find the \texttt{OralContribution} ontology a helpful starting place \parencite{Somerville}.

\begin{figure*}[t]
    \centering
    \includegraphics[width=\textwidth]{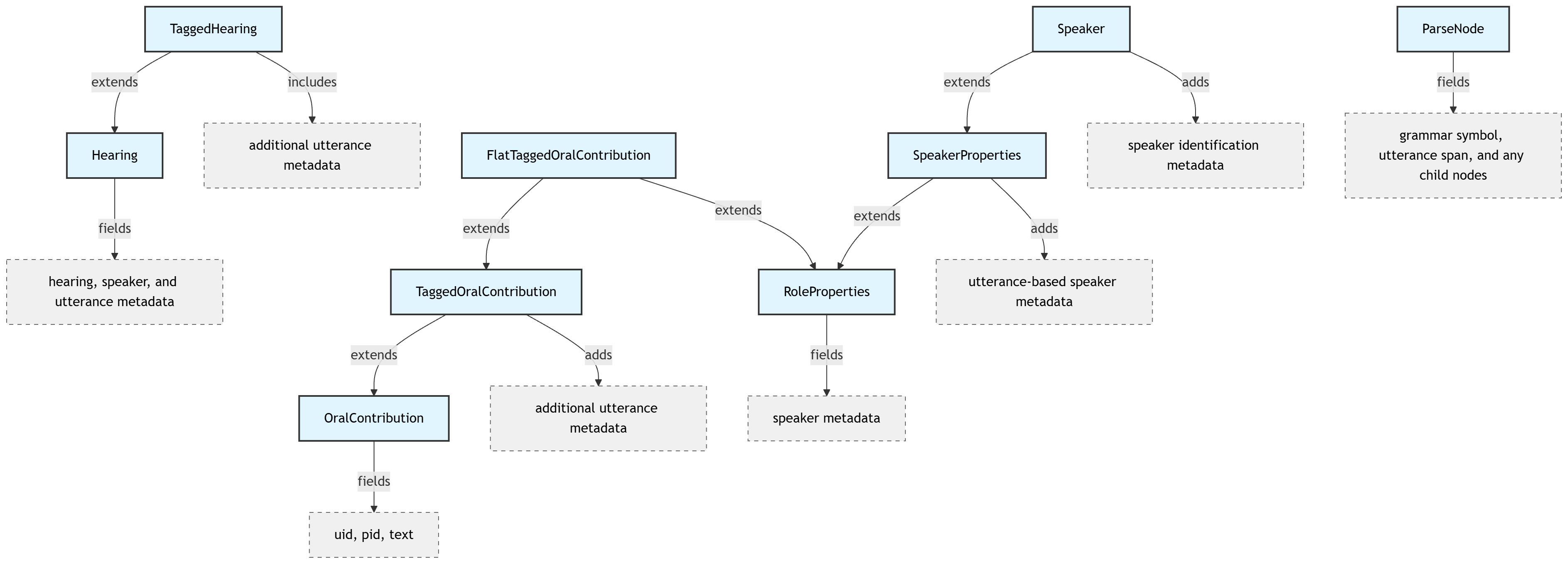}
    \caption{The final data model used for serializing, parsing, and restructuring hearing transcripts. Grey blocks denote the fields associated with different dataclasses. Blue blocks denote dataclasses. The core enumerables are defined in more depth in later sections.}
    \label{data model ontology}
\end{figure*}

This ontology was iteratively distilled and restructured based on manual dataset review coupled with SME feedback according to a shortened version of the process described in \textcite{Rudnicki2016}. We then identified features relating speaker positions and roles with their behaviors through this process. This included hearing metadata as well as features implied from different combinations of lexical and metadata properties. The final set of speaker and utterance features are organized into a coherent ontology that allows the design of inference rules for speaker role and utterance category tagging. The data model implemented for this ontology is shown in Figure \ref{data model ontology}.

\subsection{Pipeline}
\begin{figure*}[t]
    \centering
    \includegraphics[width=1.00\textwidth]{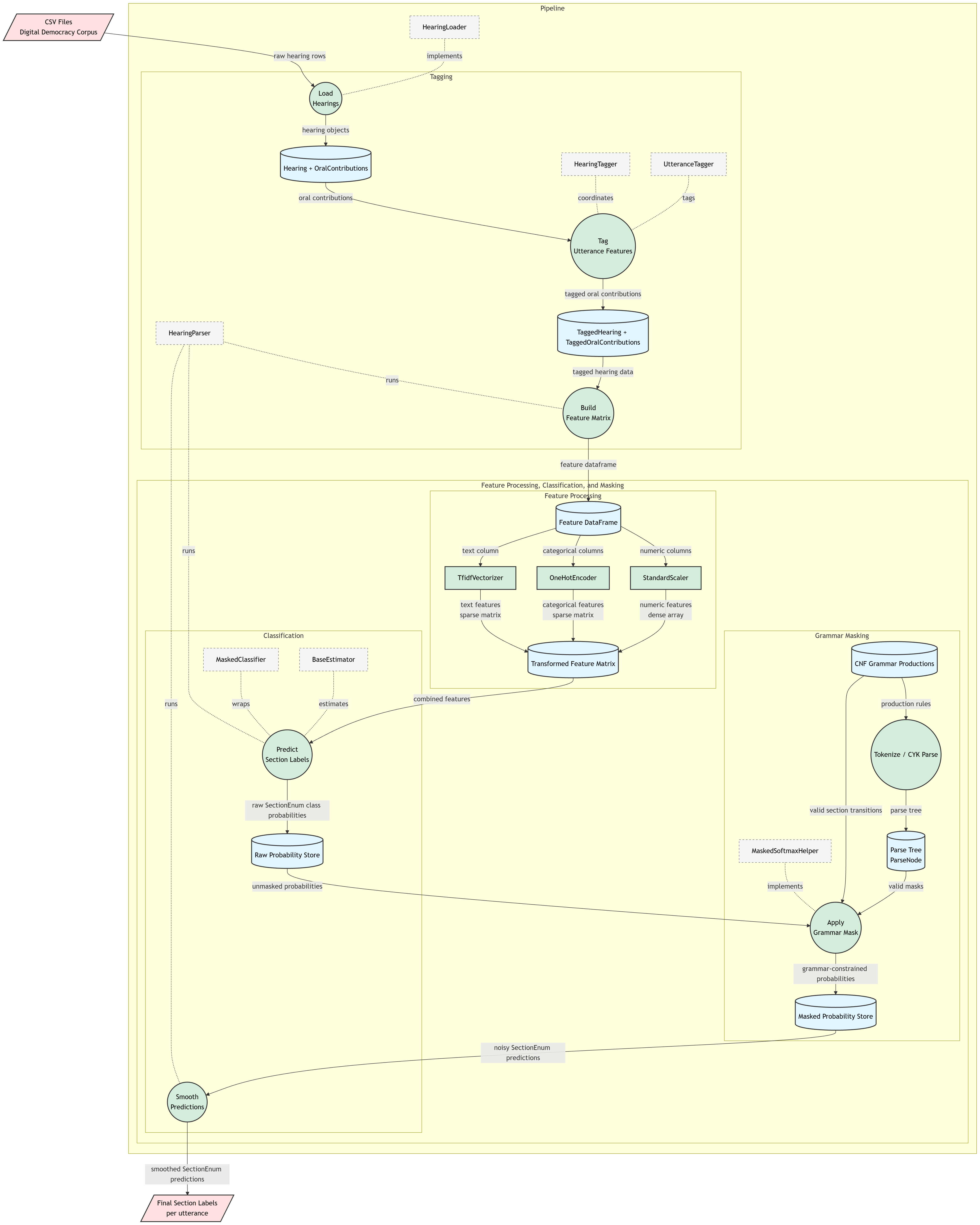}
    \caption{The pipeline used in our utterance section classifier. Green blocks shapes represent processes. Blue cylinders represent data stores. Grey blocks denote classes. Red parallelograms denote input/output data.}
    \label{pipeline}
\end{figure*}

As with any good datascience project, the bulk of our pipeline was dedicated to cleaning, serializing, and parsing our dataset. The full pipeline can be seen in Figure \ref{pipeline}, but we include a more detailed explanation of our design choices below.

\subsubsection{Tag Enrichment}
As part of our hearing loading stage, speaker and utterance metadata is linked and a variety of inference rules based on \textcite{Dimitrijevik2026} and \textcite{Das2026} are used to assign broad speaker roles. Incremental utterance cleaning is used to further disambiguate speakers' roles and add contextual features to \texttt{TaggedOralContribution} and \texttt{Speaker} instances. For example, \texttt{TaggedOralContribution} instances are enriched with as \texttt{section\_cues} and \texttt{pids\_mentioned}.

\subsubsection{Vectorization}
After disambiguating speaker roles and complex utterance tags as much as possible, hearing and speaker features are propagated down to utterances, which are flattened into rows within a \texttt{pandas.DataFrame} instance. This dataframe is passed into an \texttt{sklearn.compose.ColumnTransformer} instance, in which numerical features are standardized, categorical features are onehot-encoded, and TF-IDF embeddings are extracted from utterances' normalized texts.

These features are all passed into either a logistic regression classifier or a multi-layer perceptron (MLP) classifier. Both models are trained on a 0.75-0.25 train-test split of our ground-truth set. 

\subsubsection{Masking and Classification} \label{masking section}
In parallel with the vectorization process above, utterances (\texttt{TaggedOralContribution}s) are tokenized into \texttt{SpeakerPositionEnum} objects based on the position associated with their speaker. While this representation discards a significant amount of utterance feature information, it makes them more amenable to structural validation.

Using a combination of SME advice and manual tuning, we construct the Context-Free Grammar (CFG) described in \ref{grammar}. This grammar treats \texttt{SpeakerPositionEnum} objects as terminal symbols. To perform the aforementioned structural validation, we use the grammar rules and a Cocke–Younger–Kasami (CYK) parser to construct rhetorical parse trees for each tokenized utterance. Even though most utterances have at least 1 parse, factors like incorrect speaker role, bad utterance tags, and non-standard committee hearing structures can make such a strict parse impossible. As a fall-back, we use \texttt{SectionEnum} instances' reachability from utterances' associated \texttt{SpeakerPositionEnum}s to construct lists of valid section tags. Probabilities associated with invalid section tags by the classifier are set to 0 prior to final class assignment.

We evaluate this pipeline with 2 BaseEstimators: a 3-layer MLP model and a logistic regression-based classifier. The parameters for these are described by Table \ref{tab:classifier_hyperparameters}.

\begin{table}[htbp]
\centering
\caption{Classifier hyperparameters.}
\label{tab:classifier_hyperparameters}
\begin{tabular}{lll}
\hline
\textbf{Classifier} & \textbf{Parameter} & \textbf{Value} \\
\hline
Logistic Regression & \texttt{class\_weight} & \texttt{balanced} \\
Logistic Regression & \texttt{max\_iter} & \texttt{2000} \\
Logistic Regression & \texttt{solver} & \texttt{lbfgs} \\
\hline
MLP Classifier & \texttt{hidden\_layer\_sizes} & \texttt{(10, 10)} \\
MLP Classifier & \texttt{max\_iter} & \texttt{2000} \\
MLP Classifier & \texttt{solver} & \texttt{adam} \\
\hline
\end{tabular}
\end{table}

\subsection{Smoothing}
Since we did not train a transition detector and our core classifiers are not sequence models, our classifier may occasionally classify utterances in close proximity to each other as vastly different sections. This is especially prevalent in cases where To mitigate this, we implement a smoothing function over our outputted utterance stage predictions when labeling a full bill discussion, ensuring that an arbitrary utterance within a stage does not match the tags from the rest of the utterances.

\section{Results}
\subsection{Parsing}
Though developed based on our ground-truth annotations, we evaluated our grammar on the broader set of  2015-2016 and 2017-2018 California senate and assembly committee hearings with the selection criteria defined in Section \ref{dataset}. We did this by running our full tagging pipeline on 7030 hearings hearings and counting the number of hearings that had at least 1 parse. From this, we successfully parsed 5252 of the 7030 hearings, or 74.71\%. Every hearing with a successful parse under our grammar had at least 1 valid parse tree. 39 of the 1778 failure cases (2.19\%) were due to tagging failures.

\subsection{Section Tagging}
To evaluate our model, we used a 75-25 train-test split of our ground-truth set at the hearing level and trained two separate models.
The confusion matrices associated with the MLP and logistic regression classifiers can be seen in \ref{fig:confusion_matrix}. While not pictured, it's worth noting that a few other MLP architectures were evaluated and the one compared against the logistic regressor proved to be the best. Also worth noting is the high accuracies associated with the `EXPERT\_TESTIMONY' and `VOTE' sections. This intuitively makes sense, since votes are typically conducted by the committee secretary and follow a highly structured format. Similarly, experts are often the only speakers during `EXPERT\_TESTIMONY', with only occasional remarks by committee members. In contrast, sections like the `LEGISLATOR\_DISCUSSION' and `PUBLIC\_COMMENTS' are misclassified more often due to their frequent mixing. While still low, the `VOTE' section probably has a relatively high misclassification rate as `LEGISLATOR\_DISCUSSION' for similar reasons. These stages are more unstructured and even with grammar-restricted `attention', they tend to vary significantly more.

\begin{figure*}[t]
    \centering
    \includegraphics[width=0.75\textwidth]{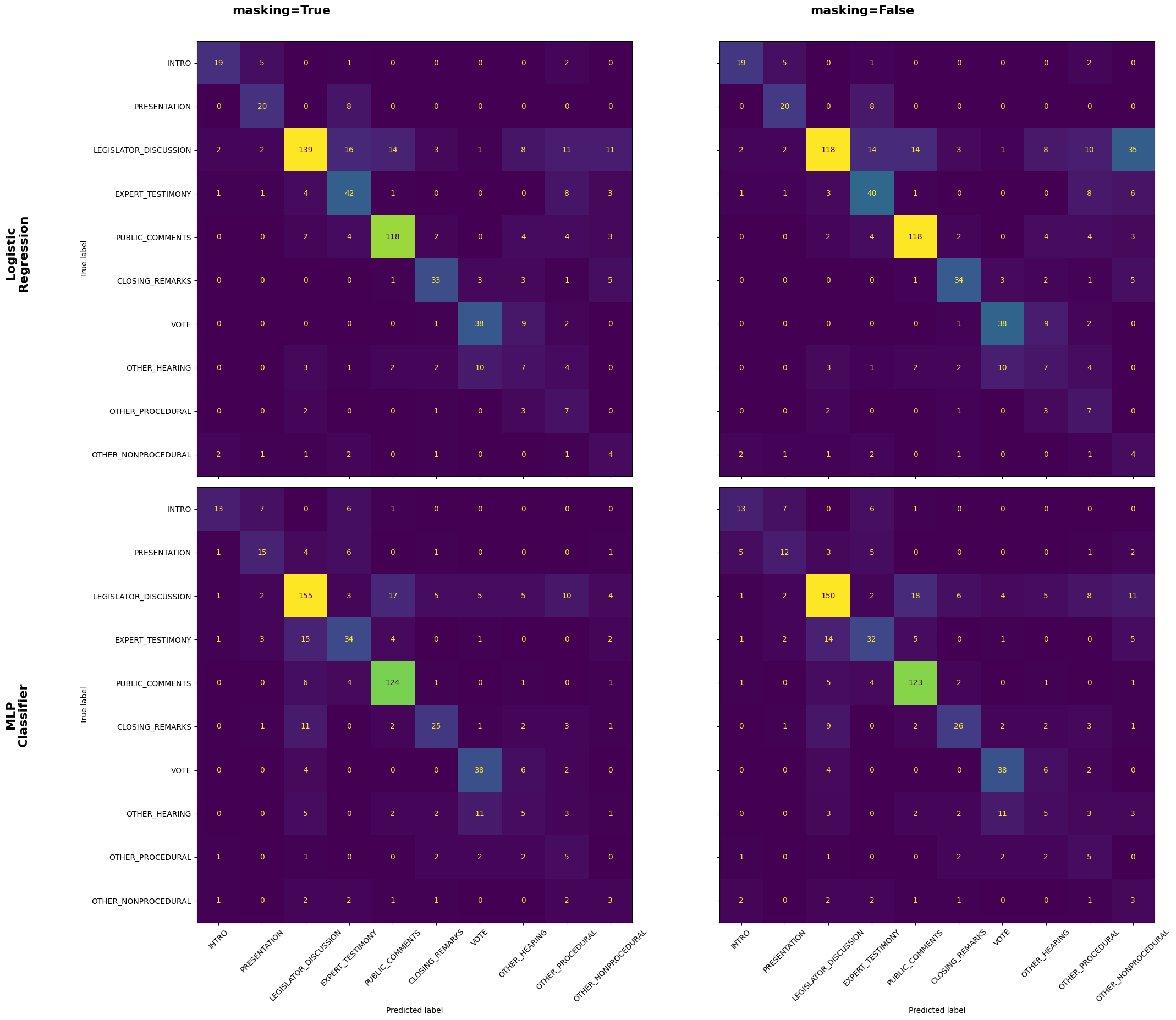}
    \caption{The confusion matrix of our utterance classifier. Our classifier categorizes each utterance into one of ten possible categories: author introduction, bill presentation, legislator discussion, expert testimony, public comments, closing remarks, voting, and 3 fallback sections. These confusion matrices are based on evaluation on a validation set of bill discussions randomly selected from the manually labeled ground truth samples.}
    \label{fig:confusion_matrix}
\end{figure*}

The table of our classifier evaluation metrics can be seen in Table \ref{tab:classifier masking comparison} and Table \ref{tab:lr class report}.  Overall, the logistic regression classifier achieves a macro average precision of 58.7\%, recall of 62.4\%, and F1-score of 59.0\% across all categories.

\begin{table}[htbp]
\centering
\caption{Classifier evaluation macro metrics with and without masking.}
\label{tab:classifier masking comparison}
\begin{tabularx}{\linewidth}{>{\raggedright\arraybackslash}p{0.1\linewidth}lrrrr}
\hline
\textbf{Model} & \textbf{Mask} & \textbf{Accuracy} & \textbf{Precision} & \textbf{Recall} & \textbf{F1} \\
\hline
LR & True  & 0.701 & 0.587 & 0.624 & 0.590 \\
LR & False & 0.665 & 0.581 & 0.613 & 0.575 \\
MLP & True  & 0.685 & 0.546 & 0.535 & 0.533 \\
MLP & False & 0.668 & 0.516 & 0.520 & 0.510 \\
\hline
\end{tabularx}
\end{table}

\begin{table}[htbp]
\centering
\caption{Logistic regression classifier evaluation metrics. For readability purposes, section names have been replaced with their corresponding grammar symbols as described in \ref{tab:enum_symbols}.}
\label{tab:lr class report}
\begin{tabularx}{\linewidth}{>{\raggedright\arraybackslash}p{0.2\linewidth}rrrr}
\hline
\textbf{Class} & \textbf{Precision} & \textbf{Recall} & \textbf{F1-score} & \textbf{Support} \\
\hline
$S_0$ & 0.792 & 0.704 & 0.745 & 27  \\
$S_1$ & 0.690 & 0.714 & 0.702 & 28  \\
$S_2$ & 0.921 & 0.671 & 0.777 & 207 \\
$S_3$ & 0.568 & 0.700 & 0.627 & 60  \\
$S_4$ & 0.868 & 0.861 & 0.864 & 137 \\
$S_5$ & 0.767 & 0.717 & 0.742 & 46  \\
$S_6$ & 0.731 & 0.760 & 0.745 & 50  \\
$S_7$ & 0.206 & 0.241 & 0.222 & 29  \\
$S_8$ & 0.175 & 0.538 & 0.264 & 13  \\
$S_9$ & 0.154 & 0.333 & 0.211 & 12  \\
\hline
\textbf{Accuracy}              &       &       & 0.701 & 609 \\
\textbf{Macro avg}             & 0.587 & 0.624 & 0.590 & 609 \\
\textbf{Weighted avg}          & 0.765 & 0.701 & 0.723 & 609 \\
\hline
\end{tabularx}
\end{table}

\subsection{Failure Case Analysis}
\label{sec:failure-analysis}

Failure-exclusive \texttt{SpeakerPositionEnum} ngrams reveal several recurring mismatches between observed speaker sequences and our current grammar. Brief secretarial or procedural interruptions can occur within expert testimony even though the \texttt{SECRETARY} role isn't allowed by the grammar described in \ref{grammar}. Similarly, the inclusion of expert and bill author commentary within  \texttt{PUBLIC\_COMMENTS} sections indicates a certain level of discourse overlap and annotation ambiguity that is not fully captured by the current model or grammar.
This would therefore see significant improvement from additional feature information or grammar granularity.

Failures associated with \texttt{SpeakerPositionEnum} sequences in the \texttt{OTHER\_HEARING} sections also suggest that administrative exchanges are underrepresented. In particular, repeated, consecutive \texttt{SECRETARY} turns are not captured by the current \texttt{OTHER\_HEARING} production.

\texttt{SpeakerPositionEnum} sequences associated with \texttt{LEGISLATOR\_DISCUSSION} and \texttt{VOTE} are usually compatible with their respective productions, suggesting that the actual parse failures may arise higher-level ordering constraints. This is especially relevant for \texttt{VOTE} sections, where repeated alternation between \texttt{SECRETARY} and \texttt{COMMITTEE\_MEMBER} speakers is consistent with ordinary roll-call procedures but is disproportionately associated with failed parses.

Overall, the failure cases suggest two main limitations:
locally unsupported speaker roles within otherwise valid sections,
and overly restrictive higher-level section production rules.

These results suggest looser higher-level production rules might offer better parses, while more granular and nuanced local section production rules are required. This seems to align with the idea that there are subsections with more granular role requirements within the current discourse sections. These failures also suggest that there may be greater variability in the \texttt{SpeakerPositionEnum} allowed within each of these sections than between them.


\section{Conclusion}
In this paper, we presented a novel approach to parsing and modeling legislative discourse. This proved very useful and enabled the construction of a logistic classifier that attained a macro F1 score of 59.0\% on a manually curated ground-truth set of utterance discourse stage labels.

There is still significant work to be done in this field. Ultimately, a grounded approach to discourse modeling holds significant promise for engaging the public with legislative policy, decision-making, and democratic accountability.
This should improve performance in downstream tasks support Democratic accountability including legislator, contradiction and inconsistency detection, implicit bias and misrepresentation tracking and policy drifted over time.

\section{Future Work}
While significant progress was made towards the complete annotation of the discourse structures associated with different legislative hearing, there is still lots of potential for future work.

Future work should focus on serializing and enumerating Robert's Rules of Order (RROO) phrases common to many legislative discussions. The more detailed motion and speaker position information associated with RROO rules' designated usage could be used to identify transitional speech acts within transcripts and further segment the transcripts according to the \texttt{SpeechActEnum} and \texttt{Motion} parts of our ontology.

Additional keywords and key phrases could be extracted based on the correlation of n-grams' TF-IDF embeddings with known utterance, speaker, and section tags. These keyphrases would support a more interpretable and extensible representation. This would allow for even more detailed discourse grammar rules as well as more complex feature extraction.

Due to its restrictiveness, the grammar described in Appendix \ref{grammar} failed to generate rhetorical parse trees for all utterances. This limited its effectiveness in masking our classifiers' outputs and may be the cause behind a significant portion of the failure cases, as noted in Section \ref{masking section}. Future work could remedy this by inducing a probabilistic context-free grammar on the subset of annotated utterances.
\begin{figure*}[t]
    \centering
    \includegraphics[width=\textwidth]{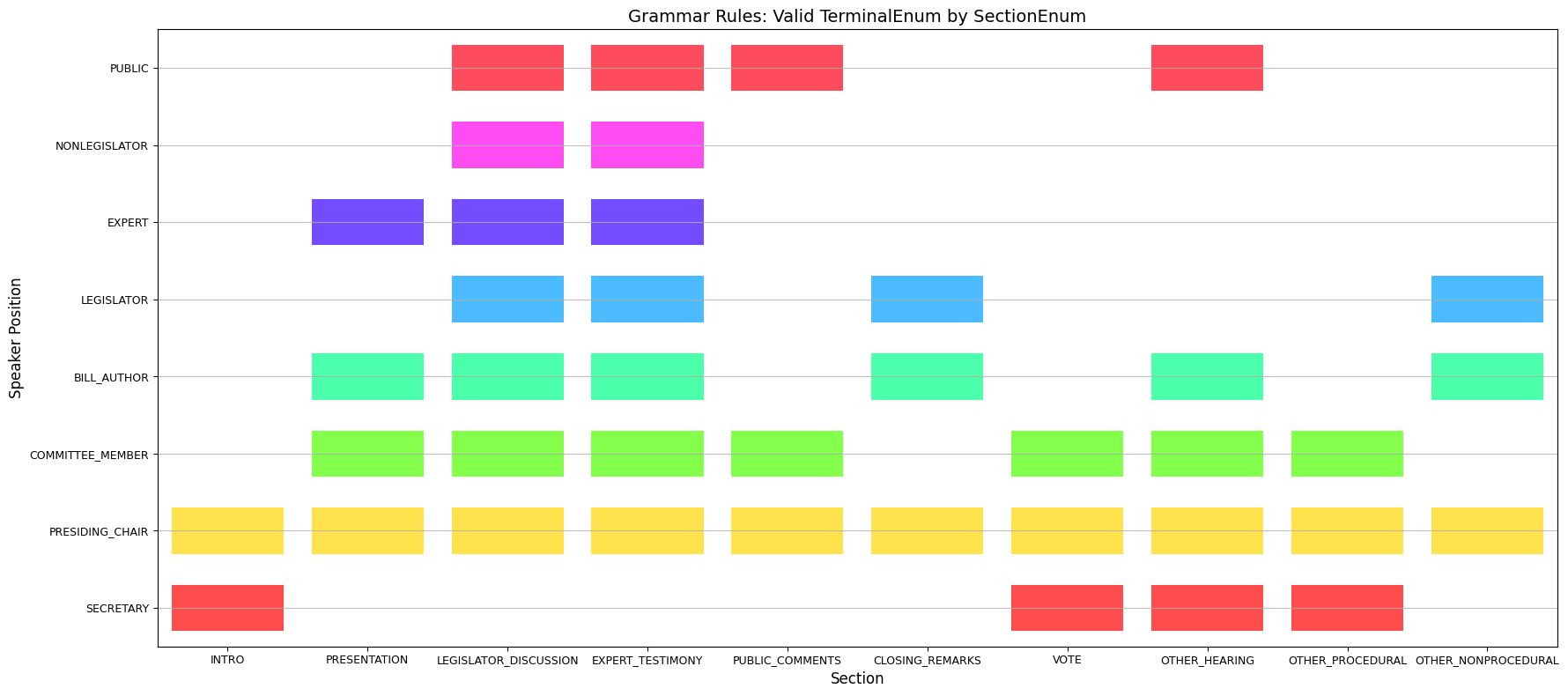}
    \caption{Speaker roles and positions associated with different sections. Voting subsections' speaker role requirements may vary, but they represent subsets of the roles attached in the diagram. These represent the section-terminal relations given in Appendix \ref{grammar}}
    \label{role associations}
\end{figure*}

Lastly, we recommend further testing with more advanced classifiers. Due to time constraints, we were only able to compare a 3-layer MLP model and a logistic regression-based classifier. But if work by \textcite{Wang2018} and \textcite{Knaebel2021} is anything to go by, there are still many promising architectures to be tried.

\section{Acknowledgements}
ChatGPT 5.5 and Claude Sonnet 4.5 were used to implement portions of various tokenization, tagging, masking, and classification classes.
ChatGPT 5.5 was used to implement the visualizations seen in figures \ref{speaker_section_frequencies}, \ref{t-sne clusters}, styling for \ref{fig:confusion_matrix}, \ref{role associations}, and formatting for appendices \ref{ngram correlations}, \ref{grammar}.

\printbibliography

@inproceedings{Khosmood2024,
  author    = {Khosmood, Foaad and Dekhtyar, Alexander and Ellwein, Sarah and White, Brian},
  title     = {The Digital Democracy Corpus: Comprehensive Proceedings of Four State Legislatures 2015--2018},
  booktitle = {Digital Humanities},
  address   = {Washington, DC},
  month     = aug,
  year      = {2024}
}

@conference{Dimitrijevik2026,
    author = {Sofija Dimitrijevik and Dr. Foaad Khosmood},
    booktitle = {Cal Poly Engineering Project Showcase},
    title = {Detection of Self-Introductions in Legislative Testimony},
    year = {2026}
}

@conference{Das2026,
    author = {Pallavi Das, Kasey Liu and Dr. Foaad Khosmood},
    booktitle = {Cal Poly Engineering Project Showcase},
    title = {Name Detections from Legislative Speech Self-Introductions},
    year = {2026}
}

@inproceedings{Irani,
author = {Irani, Arman and Park, Ju Yeon and Esterling, Kevin and Faloutsos, Michalis},
title = {A Discourse Analysis Framework for Legislative and Social Media Debates},
year = {2025},
isbn = {9798400714832},
publisher = {Association for Computing Machinery},
address = {New York, NY, USA},
url = {https://doi.org/10.1145/3717867.3717918},
doi = {10.1145/3717867.3717918},
booktitle = {Proceedings of the 17th ACM Web Science Conference 2025},
pages = {199–209},
numpages = {11},
location = {
},
series = {Websci '25}
}

@inproceedings{Knaebel2021,
    title = "discopy: A Neural System for Shallow Discourse Parsing",
    author = "Knaebel, Ren{\'e}",
    editor = "Braud, Chlo{\'e}  and
      Hardmeier, Christian  and
      Li, Junyi Jessy  and
      Louis, Annie  and
      Strube, Michael  and
      Zeldes, Amir",
    booktitle = "Proceedings of the 2nd Workshop on Computational Approaches to Discourse",
    month = nov,
    year = "2021",
    address = "Punta Cana, Dominican Republic and Online",
    publisher = "Association for Computational Linguistics",
    url = "https://aclanthology.org/2021.codi-main.12/",
    doi = "10.18653/v1/2021.codi-main.12",
    pages = "128--133",
}

@article{Li2022,
	title = {A survey of discourse parsing},
	volume = {16},
	issn = {2095-2236},
	url = {https://doi.org/10.1007/s11704-021-0500-z},
	doi = {10.1007/s11704-021-0500-z},
	number = {5},
	journal = {Frontiers of Computer Science},
	author = {Li, Jiaqi and Liu, Ming and Qin, Bing and Liu, Ting},
	month = jan,
	year = {2022},
	pages = {165329},
}

@article{Miltsakaki2004,
author = {Miltsakaki, Eleni and Prasad, Rashmi and Joshi, Aravind and Webber, Bonnie},
year = {2004},
month = {04},
pages = {},
title = {The Penn Discourse Treebank},
url = {https://www.researchgate.net/publication/2903359_The_Penn_Discourse_Treebank}
}

@misc{Plural2023,
  author       = {{Plural}},
  title        = {{About Open States}},
  year         = {2023},
  howpublished = {\url{https://open.pluralpolicy.com/about/}},
  note         = {Open States project description and history. Accessed: 2026-06-28}
}

@book{Rudnicki2016,
	address = {Gaithersburg, MD},
	author = {Ron Rudnicki and Barry Smith and Tanya Malyuta and William Mandrick},
	editor = {},
	publisher = {NIST},
	title = {Best Practices of Ontology Development (Nist White Paper)},
	year = {2016},
    url = {"https://www.nist.gov/system/files/documents/2021/10/14/nist-ai-rfi-cubrc_inc_002.pdf"}
}

@online{Smethurst2016,
  author  = {Smethurst, Michael},
  title   = {Modelling Parliament(s?)},
  date    = {2016-09-09},
  organization = {Parliamentary Digital Service},
  url     = {https://pds.blog.parliament.uk/2016/09/09/modelling-parliaments/},
  urldate = {2026-05-26}
}

@online{Somerville,
  author  = {{UK Parliament}},
  editor  = {Somerville, Anya},
  title   = {UK Parliament Ontologies},
  organization = {House of Commons Library},
  url     = {https://ukparliament.github.io/ontologies/},
  urldate = {2026-05-26}
}

@misc{Turk2017,
  author       = {Turk, James et. al.},
  title        = {{Open States API v2: Root Nodes}},
  year         = {2017},
  month        = jan,
  day          = {9},
  url          = {{https://docs.openstates.org/data/categorization/}},
  note         = {Project started January 9, 2017.}
}

@inproceedings{Wang2018,
    title = "Toward Fast and Accurate Neural Discourse Segmentation",
    author = "Wang, Yizhong  and
      Li, Sujian  and
      Yang, Jingfeng",
    editor = "Riloff, Ellen  and
      Chiang, David  and
      Hockenmaier, Julia  and
      Tsujii, Jun{'}ichi",
    booktitle = "Proceedings of the 2018 Conference on Empirical Methods in Natural Language Processing",
    month = oct # "-" # nov,
    year = "2018",
    address = "Brussels, Belgium",
    publisher = "Association for Computational Linguistics",
    url = "https://aclanthology.org/D18-1116/",
    doi = "10.18653/v1/D18-1116",
    pages = "962--967",
}

\FloatBarrier
\appendix

\section{Top-10 Most Highly Correlated N-Grams}\label{ngram correlations}

\begin{sidewaystable}[htbp]
\centering
\caption{Top 10 n-grams most closely associated with each \texttt{SpeakerPositionEnum}, based on Pearson's $r$ in a one-vs-rest comparison.}
\label{speaker_ngrams}
\small
\setlength{\tabcolsep}{3pt}
\renewcommand{\arraystretch}{1.15}
\begin{tabularx}{\textheight}{
    >{\centering\arraybackslash}p{0.035\textheight}
    >{\raggedright\arraybackslash}X
    >{\raggedright\arraybackslash}X
    >{\raggedright\arraybackslash}X
    >{\raggedright\arraybackslash}X
    >{\raggedright\arraybackslash}X
    >{\raggedright\arraybackslash}X
}
\hline
\textbf{Rank} &
\textbf{\texttt{PUBLIC}} &
\textbf{\texttt{EXPERT}} &
\textbf{\texttt{BILL\_AUTHOR}} &
\textbf{\texttt{COMMITTEE\_MEMBER}} &
\textbf{\texttt{PRESIDING\_CHAIR}} &
\textbf{\texttt{SECRETARY}} \\
\hline

1 &
of \texttt{ORG} &
that are &
aye vote &
so i m &
witnesses in &
\texttt{PRESIDING\_CHAIR} aye \\
\hline

2 &
\texttt{ORG} in &
the may &
for your &
move \texttt{BILL} &
a motion &
aye \texttt{COMMITTEE\_MEMBER} aye \\
\hline

3 &
\texttt{ORG} in support &
in the &
with me &
i m just &
we ll &
aye \texttt{ORG} aye \\
\hline

4 &
in support &
the governor s &
with me today &
I know &
seeing none &
\texttt{ORG} aye \\
\hline

5 &
\texttt{NONLEGISLATOR} on behalf of \texttt{ORG} &
the governor &
an aye vote &
I was &
any other &
aye \texttt{COMMITTEE\_MEMBER} aye \texttt{COMMITTEE\_MEMBER} aye \\
\hline

6 &
with \texttt{ORG} &
that we &
I respectfully ask for your aye vote &
I do &
call the roll &
\texttt{GPE} aye \\
\hline

7 &
\texttt{NONLEGISLATOR} with \texttt{ORG} &
with our &
ask for an aye vote &
kind of &
in opposition &
aye lackey \\
\hline

8 &
my name is \texttt{NONLEGISLATOR} &
of those &
to present \texttt{BILL} &
it s going to &
witnesses in opposition &
\texttt{PRESIDING\_CHAIR} aye \texttt{PRESIDING\_CHAIR} aye \\
\hline

9 &
in strong support &
years ago &
mr chair &
I don t &
witnesses in support &
aye monning \\
\hline

10 &
mr chair and members \texttt{NONLEGISLATOR} &
of the &
I have \texttt{NONLEGISLATOR} &
going to &
thank you &
monning aye \\

\hline
\end{tabularx}
\end{sidewaystable}

\begin{sidewaystable}[htbp]
\centering
\caption{Top 10 n-grams most closely associated with each \texttt{SectionEnum}, based on Pearson's $r$ in a one-vs-rest comparison. For readability purposes, column headers have been replaced with their corresponding grammar symbols as described in \ref{tab:enum_symbols}.}
\label{section_ngrams}
\scriptsize
\renewcommand{\arraystretch}{1.15}
\begin{tabularx}{\textheight}{
    >{\centering\arraybackslash}p{0.025\textheight}
    >{\raggedright\arraybackslash}X
    >{\raggedright\arraybackslash}X
    >{\raggedright\arraybackslash}X
    >{\raggedright\arraybackslash}X
    >{\raggedright\arraybackslash}X
    >{\raggedright\arraybackslash}X
    >{\raggedright\arraybackslash}X
    >{\raggedright\arraybackslash}X
    >{\raggedright\arraybackslash}X
    >{\raggedright\arraybackslash}X
}\hline
\textbf{Rank} &
$\mathbf{S_0}$ &
$\mathbf{S_1}$ &
$\mathbf{S_2}$ &
$\mathbf{S_3}$ &
$\mathbf{S_4}$ &
$\mathbf{S_5}$ &
$\mathbf{S_6}$ &
$\mathbf{S_7}$ &
$\mathbf{S_8}$ &
$\mathbf{S_9}$ \\
\hline

1 &
whenever you re ready &
with me today &
I think &
two minutes &
in support &
aye vote &
call the roll &
to zero that bill is out &
a quorum &
thank you mr chair thank you \\
\hline

2 &
\texttt{BILL\_AUTHOR} welcome &
with me &
don t &
to support \texttt{BILL} and &
\texttt{ORG} in &
would you like to close &
do pass &
to zero &
call on &
chair thank you members \\
\hline

3 &
enjoys a do pass recommendation &
to present \texttt{BILL} &
so I &
thank you for your time &
of \texttt{ORG} &
for your &
\texttt{PRESIDING\_CHAIR} aye &
voting aye &
\texttt{PRESIDING\_CHAIR} here &
mr chair thank you members \\
\hline

4 &
is \texttt{BILL} &
I have \texttt{NONLEGISLATOR} &
that s &
benefit from &
\texttt{ORG} in support &
I respectfully ask for your aye vote &
please call the roll &
18 to zero &
so let s &
thank you mr chair thank you members \\
\hline

5 &
mr \texttt{BILL\_AUTHOR} &
today I have &
I don t &
your first &
\texttt{NONLEGISLATOR} on behalf of \texttt{ORG} &
you may close &
the motion is do pass &
is out file &
\texttt{COMMITTEE\_MEMBER} here &
you mr chair thank you members \\
\hline

6 &
be presenting on \texttt{BILL} &
\texttt{ORG} to &
it s &
we are the &
in opposition &
an aye vote &
aye \texttt{COMMITTEE\_MEMBER} aye &
out file &
establish a quorum &
I also want to \\
\hline

7 &
\texttt{BILL} which is &
to testify in support &
is that &
my name is \texttt{NONLEGISLATOR} and i m &
with \texttt{ORG} &
like to &
the roll open &
18 to &
\texttt{ORG} here &
I also want to thank \\
\hline

8 &
file item 4 &
today to &
that you &
found that &
\texttt{NONLEGISLATOR} with \texttt{ORG} &
ask for an aye vote &
aye lackey &
holden mccarty &
please take &
thank you all \\
\hline

9 &
file item 4 \texttt{BILL} &
with me today to &
but I &
very well &
witnesses in &
comments seeing none &
\texttt{ORG} aye &
holden mccarty aye &
a motion by ms reyes &
good to \\
\hline

10 &
\texttt{BILL} it enjoys &
to present \texttt{BILL} which &
trying to &
at \texttt{ORG} i &
in strong support &
a motion &
\texttt{GPE} aye &
holden mccarty aye mccarty &
by ms reyes &
also want to thank \\

\hline
\end{tabularx}
\end{sidewaystable}

\FloatBarrier
\section{Discourse Grammar}\label{grammar}

\subsection{Enumerables and Grammar Symbols}
\begin{table}[htbp]
\centering
\caption{Top-level, section, and speaker enumerables and their corresponding grammar symbols.}
\label{tab:enum_symbols}
\begin{tabular}{lll}
\hline
\textbf{Enum Symbol Type} & \textbf{Symbol Name} & \textbf{Notation} \\
\hline
\texttt{TOP} & \texttt{START}           & $A$ \\
\texttt{TOP} & \texttt{MIDDLE}          & $B$ \\
\texttt{TOP} & \texttt{UPPER\_MIDDLE}   & $B_{1}$ \\
\texttt{TOP} & \texttt{LOWER\_MIDDLE}   & $B_{2}$ \\
\texttt{TOP} & \texttt{END}             & $C$ \\
\hline
\texttt{SectionEnum} & \texttt{INTRO}                 & $S_{0}$ \\
\texttt{SectionEnum} & \texttt{PRESENTATION}          & $S_{1}$ \\
\texttt{SectionEnum} & \texttt{LEGISLATOR\_DISCUSSION}& $S_{2}$ \\
\texttt{SectionEnum} & \texttt{EXPERT\_TESTIMONY}     & $S_{3}$ \\
\texttt{SectionEnum} & \texttt{PUBLIC\_COMMENTS}       & $S_{4}$ \\
\texttt{SectionEnum} & \texttt{CLOSING\_REMARKS}      & $S_{5}$ \\
\texttt{SectionEnum} & \texttt{VOTE}                  & $S_{6}$ \\
\texttt{SectionEnum} & \texttt{OTHER\_HEARING}        & $S_{7}$ \\
\texttt{SectionEnum} & \texttt{OTHER\_PROCEDURAL}     & $S_{8}$ \\
\texttt{SectionEnum} & \texttt{OTHER\_NONPROCEDURAL}  & $S_{9}$ \\
\hline
\texttt{SpeakerPositionEnum} & \texttt{SECRETARY}         & $t_{9}$ \\
\texttt{SpeakerPositionEnum} & \texttt{PRESIDING\_CHAIR}  & $t_{8}$ \\
\texttt{SpeakerPositionEnum} & \texttt{CHAIRMAN}          & $t_{7}$ \\
\texttt{SpeakerPositionEnum} & \texttt{VICE\_CHAIRMAN}    & $t_{6}$ \\
\texttt{SpeakerPositionEnum} & \texttt{COMMITTEE\_MEMBER} & $t_{5}$ \\
\texttt{SpeakerPositionEnum} & \texttt{BILL\_AUTHOR}      & $t_{4}$ \\
\texttt{SpeakerPositionEnum} & \texttt{LEGISLATOR}        & $t_{3}$ \\
\texttt{SpeakerPositionEnum} & \texttt{EXPERT}            & $t_{2}$ \\
\texttt{SpeakerPositionEnum} & \texttt{NONLEGISLATOR}     & $t_{1}$ \\
\texttt{SpeakerPositionEnum} & \texttt{PUBLIC}            & $t_{0}$ \\
\hline
\end{tabular}
\end{table}

\subsection{Production rules}
The following grammar is stored in Chomsky Normal Form (CNF) within our implementation, but is described in a looser form below for readability.
\begin{scriptsize}
\begin{align*}
\langle ROOT \rangle
    &::= \langle ROOT2 \rangle\,\langle C \rangle
\\[0.5em]
\langle ROOT2 \rangle
    &::= \langle A \rangle\,\langle B \rangle
\\[0.5em]
\langle A \rangle
    &::= \langle S_{0} \rangle\,\langle S_{1} \rangle
    \mid \langle S_{0} \rangle
    \mid \langle S_{1} \rangle
\\[0.5em]
\langle B \rangle
    &::= \langle B_{1} \rangle\,\langle B_{2} \rangle
    \mid \langle B_{1} \rangle
    \mid \langle B_{2} \rangle
    \mid \langle S_{3} \rangle\,\langle S_{2} \rangle
    \mid \langle S_{4} \rangle\,\langle S_{2} \rangle
    \mid \langle S_{3} \rangle
    \mid \langle S_{4} \rangle
    \mid \epsilon
\\[0.5em]
\langle B_{1} \rangle
    &::= \langle B_{1} \rangle\,\langle B_{1} \rangle
    \mid \langle B_{1} \rangle\,\langle S_{3} \rangle
    \mid \langle B_{1} \rangle\,\langle S_{2} \rangle
    \mid \langle S_{3} \rangle\,\langle S_{3} \rangle
\\[0.5em]
\langle B_{2} \rangle
    &::= \langle B_{2} \rangle\,\langle B_{2} \rangle
    \mid \langle B_{2} \rangle\,\langle S_{4} \rangle
    \mid \langle B_{2} \rangle\,\langle S_{2} \rangle
    \mid \langle S_{4} \rangle\,\langle S_{4} \rangle
\\[0.5em]
\langle C \rangle
    &::= \langle S_{5} \rangle\,\langle S_{6} \rangle
    \mid \langle S_{5} \rangle
    \mid \langle S_{6} \rangle
    \mid \langle C \rangle\,\langle S_{7} \rangle
\\[0.5em]
\langle S_{0} \rangle
    &::= \langle S_{8} \rangle\,\langle S_{0} \rangle
    \mid \langle S_{0} \rangle\,\langle S_{0} \rangle
    \mid \mathrm{t_{8}}
    \mid \mathrm{t_{9}}
\\[0.5em]
\langle S_{1} \rangle
    &::= \langle S_{1} \rangle\,\langle S_{9} \rangle
    \mid \langle S_{1} \rangle\,\langle S_{1} \rangle
    \mid \mathrm{t_{2}}\,\mathrm{t_{8}}
    \mid \mathrm{t_{4}}
    \mid \mathrm{t_{5}}
    \mid \mathrm{t_{8}}
\\[0.5em]
\langle S_{2} \rangle
    &::= \langle S_{2} \rangle\,\langle S_{2} \rangle
    \mid \mathrm{t_{4}}\,\mathrm{t_{0}}
    \mid \mathrm{t_{4}}\,\mathrm{t_{1}}
    \mid \mathrm{t_{4}}\,\mathrm{t_{2}}
    \mid \mathrm{t_{5}}\,\mathrm{t_{0}}
    \mid \mathrm{t_{5}}\,\mathrm{t_{1}}
    \mid \mathrm{t_{5}}\,\mathrm{t_{2}}
    \mid \mathrm{t_{8}}\,\mathrm{t_{0}}
    \mid \mathrm{t_{8}}\,\mathrm{t_{1}}
    \mid \mathrm{t_{8}}\,\mathrm{t_{2}}
    \mid \mathrm{t_{4}}
    \mid \mathrm{t_{3}}
    \mid \mathrm{t_{5}}
    \mid \mathrm{t_{8}}
\\[0.5em]
\langle S_{3} \rangle
    &::= \langle S_{3} \rangle\,\langle S_{3} \rangle
    \mid \mathrm{t_{4}}\,\mathrm{t_{2}}
    \mid \mathrm{t_{8}}\,\mathrm{t_{0}}
    \mid \mathrm{t_{1}}
    \mid \mathrm{t_{2}}
    \mid \mathrm{t_{3}}
    \mid \mathrm{t_{5}}
    \mid \mathrm{t_{8}}
\\[0.5em]
\langle S_{4} \rangle
    &::= \langle S_{4} \rangle\,\langle S_{4} \rangle
    \mid \mathrm{t_{0}}
    \mid \mathrm{t_{5}}
    \mid \mathrm{t_{8}}
\\[0.5em]
\langle S_{5} \rangle
    &::= \langle S_{5} \rangle\,\langle S_{9} \rangle
    \mid \langle S_{5} \rangle\,\langle S_{5} \rangle
    \mid \mathrm{t_{8}}\,\mathrm{t_{3}}
    \mid \mathrm{t_{8}}\,\mathrm{t_{4}}
    \mid \mathrm{t_{8}}
\\[0.5em]
\langle S_{6} \rangle
    &::= \langle S_{6} \rangle\,\langle S_{9} \rangle
    \mid \langle S_{6} \rangle\,\langle S_{6} \rangle
    \mid \mathrm{t_{8}}\,\mathrm{t_{9}}
    \mid \mathrm{t_{5}}\,\mathrm{t_{9}}
    \mid \mathrm{t_{9}}
\\[0.5em]
\langle S_{7} \rangle
    &::= \langle S_{7} \rangle\,\langle S_{7} \rangle
    \mid \mathrm{t_{0}}\,\mathrm{t_{8}}
    \mid \mathrm{t_{0}}\,\mathrm{t_{9}}
    \mid \mathrm{t_{8}}\,\mathrm{t_{0}}
    \mid \mathrm{t_{9}}\,\mathrm{t_{0}}
    \mid \mathrm{t_{9}}\,\mathrm{t_{4}}
    \mid \mathrm{t_{5}}
    \mid \mathrm{t_{8}}
\\[0.5em]
\langle S_{8} \rangle
    &::= \langle S_{8} \rangle\,\langle S_{8} \rangle
    \mid \mathrm{t_{5}}\,\mathrm{t_{9}}
    \mid \mathrm{t_{8}}
    \mid \mathrm{t_{9}}
\\[0.5em]
\langle S_{9} \rangle
    &::= \langle S_{9} \rangle\,\langle S_{9} \rangle
    \mid \mathrm{t_{8}}\,\mathrm{t_{4}}
    \mid \mathrm{t_{3}}
\end{align*}
\end{scriptsize}

\end{document}